# GoDP: Globally Optimized Dual Pathway deep network architecture for facial landmark localization in-the-wild

Yuhang Wu, Shishir K. Shah, Ioannis A. Kakadiaris[1]

{ywu35,sshah,ikakadia}@central.uh.edu

*Computational Biomedicine Lab*
*Department of Computer Science, University of Houston*
*4849 Calhoun Road, Houston, TX, 77004*

## Abstract

Facial landmark localization is a fundamental module for pose-invariant face recognition. The most common approach for facial landmark detection is cascaded regression, which is composed of two steps: feature extraction and facial shape regression. Recent methods employ deep convolutional networks to extract robust features for each step, while the whole system could be regarded as a deep cascaded regression architecture. In this work, instead of employing a deep regression network, a Globally Optimized Dual-Pathway (GoDP) deep architecture is proposed to identify the target pixels through solving a cascaded pixel labeling problem without resorting to high-level inference models or complex stacked architecture. The proposed end-to-end system relies on distance-aware softmax functions and dual-pathway proposal-refinement architecture. Results show that it outperforms the state-of-the-art cascaded regression-based methods on multiple in-the-wild face alignment databases. The model achieves 1.84 normalized mean error (NME) on the AFLW database [1], which outperforms 3DDFA [2] by 61.8%. Experiments on face identification demonstrate that GoDP, coupled with DPM-headhunter [3], is able to improve rank-1 identification rate by 44.2% compare to Dlib [4] toolbox on a challenging database.

---

[1]Corresponding author

## 1. Introduction

Facial landmark detection is the problem of localizing sali-ent facial landmarks (*e.g.*, eye corners, nose tip) on the human face. If there are $L$ facial landmarks whose locations need to be estimated, then the target space of this problem is a $2L \times 1$ feature vector where each attribute corresponds to the horizontal or vertical coordinate of a specific landmark in a given image. In biometric research, facial landmark detection is a fundamental module in many face recognition systems [5, 6, 7, 8, 9, 10]. These systems rely on landmarks to construct semantic correspondences between facial images. In real-world deployment, facial landmark detection may become a bottleneck that constrains face recognition systems to achieve optimum performance. Even though the boundaries of landmark detection have been consistently pushed forward in past years, localizing landmarks under unconstrained conditions (detecting landmarks under large head pose deviations, facial expression variations, illumination changes, and face occlusions) still remains challenging.

Common approaches for facial key-point localization are based on cascaded regression [11, 12, 2, 13]. In recent works [14, 15, 16], a confidence-map is generated for each landmark to indicate the possibility of a landmark appearing at a specific location in the original image. The prediction is made by selecting the location that has the maximum response in the confidence-map. Compared to the previous dominant approaches (e.g., cascaded regression), this winner-take-all strategy helps to suppress false alarms generated by noisy regions, improves the robustness of the algorithm, and is highly accurate in localizing visible landmarks under arbitrary head poses. This paper further discloses that the design of network strongly affects the network's performance. The deep representation generated at the bottom layer, while highly discriminative and robust for object/face representation, fails to retain enough spatial resolution after many pooling and convolutional layers. As a result, it is hard to tackle pixel-level localization/classification tasks well. This phenomenon was recently named in the image segmentation field as spatial-semantic uncertainty [17]. This limitation may potentially impacts facial landmark localization, since in many of the previous works [18, 19, 2, 13], landmark estimations are solely rely on the deep representations gener-ated by the latest layers of the networks.

To tackle the aforementioned challenges in deep network design, a globally opti-mized dual pathway (GoDP) network is proposed. In this network, all inferences are conducted on 2D score maps to facilitate gradient back-propagation. Because there are very few landmark locations activated on the 2D score maps, a distance-aware softmax function (DSL) is proposed, which reduces the false alarms in the 2D score maps. To solve the spatial-semantic uncertainty problem of deep architecture, a dual pathway model is proposed where shallow and deep layers of the network are jointly forced to maximize the possibility of a highly specific candidate region. As a result, GoDP achieved state-of-the-art performance on multiple challenging databases. To further demonstrate the contribution of the model, GoDP is embedded into a 3D-aided face recognition system. Results demonstrated that GoDP significantly improves the qual-ity of 3D face frontalization in both near-frontal and challenging poses when coupled with an automatic face detector, compared with using that Dlib [4] package (a popu-lar landmark detector used in recent face identification researches [5, 6, 7]) to localize landmarks.

The key contributions include:

- A deep network architecture that is able to generate high quality 2D score maps for accurate key-points localization without stacked architecture.

- A supervised proposal-refinement architecture to discriminatively extract spatial-semantic information from the deep network.

- A new loss function designed for reducing false alarms in the 2D score maps.

The rest of this paper is organized as follows. In Section 2, the related works in deep cascaded regression and pixel-labeling are briefly reviewed. In Section 3, GoDP and three critical components of GoDP are proposed. In Section 4,GoDP is evaluated in challenging databases of face alignment and identification. Finally, the conclusion is drawn in Section 5.

## 2. Related Work

Most of the deep architectures used for face alignment are extended from the framework of cascaded regression. Sun *et al.*[18] first employed an AlexNet-like architecture to localize five fiducial points on faces. Later, Zhang *et al.* [19] proposed a multi-task framework demonstrating that a more robust landmark detector can be built through joint learning with correlated auxiliary tasks, such as head pose and facial expression, which outperformed several shallow architectures [20, 21, 22, 23]. To address facial alignment problems under arbitrary head poses, Zhu *et al.*[12] and Jourabloo *et al.*[13] employed a deformable 3D model to jointly estimate facial poses and shape coefficients. These deep cascaded regression methods outperform multiple state-of-the-art shallow structures [24, 25, 26], and achieved remarkable performance on less controlled databases such as AFLW [1] and AFW [27]. To improve facial landmark localization accuracy, some researchers [18, 28, 29, 30] have employed local deep networks to localize facial landmarks based on facial patches. These networks rely strongly on the accuracy of a global network to select correct landmark candidate regions before being deployed. Once the global network fails, the local networks cannot fully correct the accumulated errors. A common point of the previous cascaded architectures is that they require model re-initalization when switching stages/networks. As a result, the parameters of each stage/network are optimized from a greedy stage-wise perspective, which is suboptimal.

Recent works in human pose estimation [14, 15, 16] employ 2D score maps as the targets for inference. This modification enables gradients back-propagation between stages, allows 2D feedback loops, and hence delivers an end-to-end model. In this new family of methods for key-points localization, the works of Peng [16] *et al.* and Xiao [32] *et al.* rely on a DeconvNet [31] architecture to localize facial landmarks. Even though they obtain impressive results by integrating the estimation with recurrent neural networks, the quality of face alignment is intrinsically limited by the low-quality confidence map generated by the DeconvNet as shown in Fig. 1. Wei *et al.* [15] proposed the convolutional pose machine (CPM), which employs a stacked cascaded architecture to refine body key-point predictions. This cascaded structure has multi-ple subnetworks that gradually minimize the residual errors of score maps. It is an end-to-end deep model. However, the input of each subnetwork in [15] is the original image, which ends up with a heavy and redundant architecture that is difficult for small scale deployment. Bulat *et al.* [14] employs a two-stage convolutional aggrega-tion architecture where a CNN detector is trained first, then a CNN regressor is learned based on the input of both shallow-level and deep-level features of the CNN detector to further boost the accuracy of landmark localization. The two-stage architecture contains multiple convolutional layers with large filter size. In contrast, GoDP detect the landmarks through a carefully designed coarse-to-fine framework and achieves better performance.

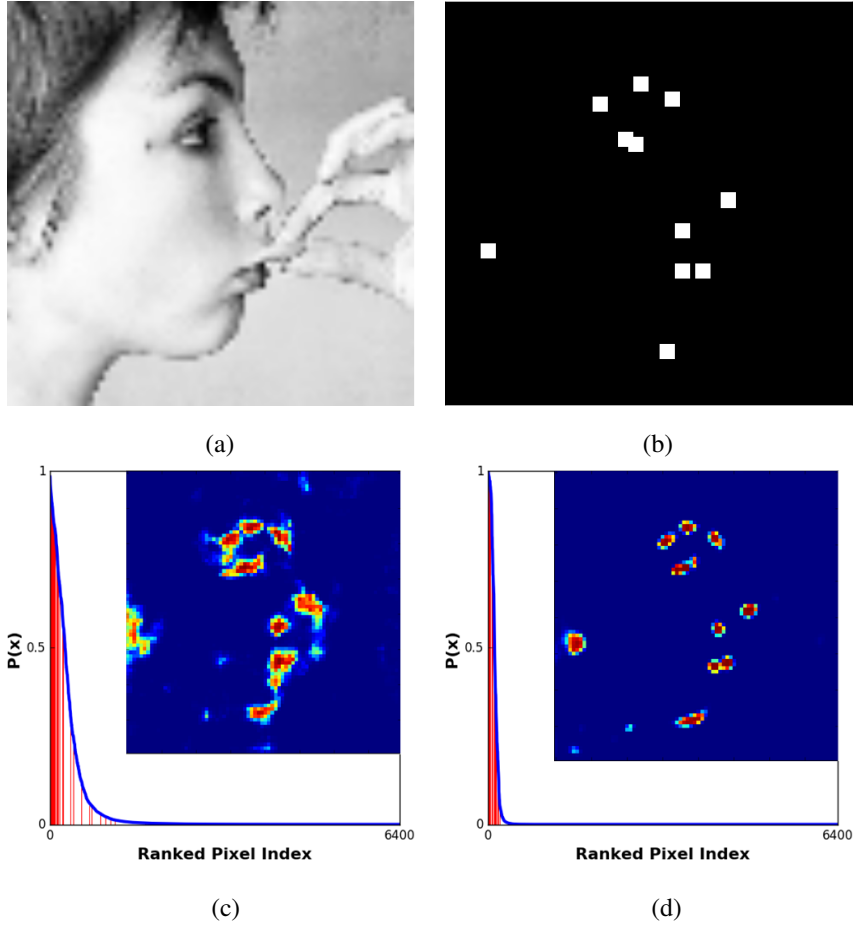

(a)                                    (b)

(c)                                    (d)

Figure 1: Score maps generated from different variations of DeconvNet. Pixels in the score maps indicate probabilities of visible facial key-points. (a) Original image, (b) Ground-truth mask, (c) DeconvNet [31], (d) GoDP. The pixel values are ranked in each score map and plotted as the blue curve line underneath. The red vertical lines indicate the pixel values in the key-point candidate positions (3×3 white patches plotted in (b)). This comparison shows that the score map generated from GoDP is clear and discriminative.

One fundamental challenge when employing 2D score maps for key-point localization is spatial-semantic uncertainty, which is critical but has not been the focus of previous works on face alignment. Ghiasi *et al.* [17] pointed out that features generated from the bottom layers of deep networks, although encoding semantic information which is robust to image and human identity variations, lack enough spatial resolution for tasks requiring pixel-level precision (*e.g.*, image segmentation, key-points localization). To tackle this problem, the authors of [16, 33] concatenated shallow-level convolutional layers to the latest convolutional layers before landmark regression. Newell *et al.* [34] proposed a heavily stacked structure by intensively aggregating shallow and deep convolutional layers to obtain better score map predictions. In image segmentation, Pinherio *et al.* [35] proposed a top-down refinement architecture which first generates robust but low-resolution score maps in a feedforward pass, then gradually refines the score maps in a top-down pass using features at lower-level layers. Although the aforementioned methods improve the deep networks' resolution for accomplishing pixel-level labeling tasks, concatenating and adding noised shallow-level features with deep-level features without regularization could be detrimental to the system's discriminative capability. Ghiasi *et al.* [17] proposed a Laplacian-pyramid-like architecture that refines the 2D score maps generated by the bottom layers by adding back features generated from top layers with the supervision of three soft-max loss layers, which provides more constraints to the refinement. In this work, a well-controlled proposal and refinement architecture is introduced for key-point localization. By imposing an appropriate supervising signal, shallow-level features in GoDP architecture are used to propose key-point candidates and deep-level features are responsible for refining the proposals and suppressing the false-alarms in the background. A new loss function is proposed to impose appropriate regularization over different network layers and guar-antee the whole architecture works as expected. Experimental results show that GoDP architecture outperforms adding-back structure employed by Newell *et al.* [34] in the DeconvNet architecture.

Recently, more works in landmark and body joint localization rely on high-level inference models such as recurrent network [16, 32], conditional random fields [36], or other graphical models [37]. GoDP does not employ high-level architecture in inference. The method improves a traditional deep network by better employing the information encoded in network layers. As a result, better and more discriminative score maps can be obtained for estimating landmark positions.

## 3. Method

In this section, deep cascaded regression is briefly reviewed, then three components of GoDP are introduced. They are the basic elements that help to address multiple challenges in 2D score map-based inference.

### 3.1. Deep cascaded regression

The intent of cascaded regression is to progressively minimize a difference $\Delta S$ between a predicted shape $\hat{S}$ and a ground-truth shape $S$ in an incremental manner. This approach contains $T$ stages, starting with an initial shape $\hat{S}^0$; the estimated shape $\hat{S}^t$ is gradually refined as:

$$\arg\min_{\mathbb{R}^t, \mathbb{F}^t} \sum_i ||\Delta S_i^t - \mathbb{R}^t(\mathbb{F}^t(\hat{S}_i^{t-1}, \mathbf{I}_i))||_2^2; \tag{1}$$

$$\hat{S}_i^t = \hat{S}_i^{t-1} + \Delta \hat{S}_i^{t-1} \tag{2}$$

where $i$ iterates over all training images. The estimated facial shape for image $\mathbf{I}_i$ in stage $t$ is denoted by $\hat{S}_i^t$; usually $\hat{S}_i^t$ can be represented as a $2L \times 1$ vector. The number of facial key-points is denoted by $L$. The function $\mathbb{F}^t(\hat{S}_i^{t-1}, \mathbf{I}_i)$ is a mapping from image space to feature space. Because the obtained features are partially determined by $\hat{S}_i^{t-1}$, these features are called 'shape-indexed features'. The function $\mathbb{R}^t(\cdot)$ is a learned mapping from feature space to target parameter space. In deep cascaded regression [18, 19, 2, 13], function $\mathbb{F}^t(\cdot)$ can be used to denote all operations before the last fully connected layer. This mapping $\mathbb{R}^t(\cdot)$ represents the operations in the last fully connected layer whose input is an arbitrary dimensions feature vector $\phi_i^t$ and output is the target parameter space.

The main problem of this architecture is that the current deep cascaded regression is greedily optimized for each stage. The learned mapping $\mathbb{R}^t$ is not end-to-end optimal with respect to the global shape increment. When training a new mapping $\mathbb{R}^t$ for stage $t$, fixing the network parameters of previous stages leads to a stage-wise suboptimal solution.

### 3.2. Optimized progressive refinement

Due to optimization problems in traditional deep cascaded architecture, a global optimization model is highly needed. However, the main difficulty in converting a cascaded regression approach into a globally optimized architecture is to back-propagate gradients between stages, where shape was usually used as a prior to initialize new cascaded stages. In GoDP, the problem is bypassed by representing landmark locations $\hat{S}_i^t$ through 2D score maps $\mathbf{\Psi}^t$ (the index $i$ is omitted for clarity), where information of landmark positions is summarized into probability values that indicate the likelihood of the existence of landmarks. Tensor $\mathbf{\Psi}^t$ denotes $(KL+1) \times W \times H$ score maps in stage $t$, where $L$ is the number of landmarks, $K$ is the number of subspaces (which will be introduced later), $W$ and $H$ are the width and height of the score maps. The extra $(KL + 1)^{th}$ channel indicates the likelihood that a pixel belongs to background. Through this representation, gradients can pass through the score maps and be back-propagated from the latest stages of cascaded model. Another insight of employing 2D probabilistic score maps is that these outputs can be aggregated and summarized with convolutional features and create feedback paths [38], which can be represented

as follows:

$$\Psi^0 = \mathbb{F}_o^0(\mathbf{I}) \tag{3}$$

$$\mathbb{F}_b^{t-1}(\mathbf{I}, \Psi^{t-1}) = \mathbb{F}_a^{t-1}(\mathbf{I}) \uplus \Psi^{t-1} \tag{4}$$

$$\Delta\Psi^{t-1} = \mathbb{F}_c^{t-1}(\mathbb{F}_b^{t-1}(\mathbf{I}, \Psi^{t-1})) \tag{5}$$

$$\Psi^t = \Psi^{t-1} + \Delta\Psi^{t-1} \tag{6}$$

where $\uplus$ denotes a layerwise concatenation in convolutional neural networks, $\mathbb{F}_o^0(\mathbf{I})$ represents the first $\Psi^0$ generated from $\mathbf{I}$ after passing through several layers in convolutional neural network, $\mathbb{F}_a^{t-1}(\cdot)$, $\mathbb{F}_b^{t-1}(\cdot)$ and $\mathbb{F}_c^{t-1}(\cdot)$ indicate different network operations with different parameter settings.

Equations 4 to 6 can be regarded as an iterative error feedback path [38]. Four of these paths are included in GoDP structure as shown in Fig. 3. Through the feedback paths, score maps generated by each stage can be directly concatenated with other convolutional layers through Eq. (4), which behaves as 'shape-indexed feature'. In contrast to [2] and [16], where score maps employed in the feedback paths are determined and synthesized through external parameters, in GoDP, $\Psi^{t-1}$ in Eq. (4) is fully determined by the parameters inside the network based on Eq. (5) and Eq. (6). Therefore, GoDP can be optimized globally.

### 3.3. 3D pose-aware score map

To model complex appearance-shape dependencies on the human face across pose variations, unlike recent works that employ a deformable shape model [13, 12], GoDP implicitly encodes 3D constraints. It is important to point out that pose is a relative concept whose actual value is susceptible to the sampling region, facial expression, and other factors. As a result, it is very hard to learn an accurate and reliable mapping from image to the pose parameters without considering fiducial points' correspondence. Instead of estimating pose parameters [39, 26, 13, 12] explicitly, pose is regarded as a general domain index that encodes multi-modality variations of facial appearance. Specifically, $K$ score maps are employed to indicate each landmark location, where $K$ corresponds to the number of partitions of head pose. In this work, the subspace partition is determined by yaw variations. For each image, one out of $K$ score maps is activated for each landmark. In this way, the network automatically encodes contextual information between appearance of landmarks under different poses. At the final stage, $K$ score maps are merged into one score map through element-wise summation.

### 3.4. Distance-aware softmax loss

Soft-max loss function has been widely used in solving pixel-labeling problems in human joint localization [15, 40], image segmentation [31, 17], and recently, facial key-point annotation [16]. One limitation of using softmax for key-point localization is that the function treats pixel-labeling as an independent classification problem, which does not take into account the distance between the labeling pixel and the ground-truth key-points. As a result, the loss function will assign equal penalty to the regions that lie very close to a key-point and also the regions on the border of an image, which should not be classified as landmark candidates. Another drawback of this loss function is that it assigns equal weight to negative and positive samples, which may lead the network to converge into a local minimum, where every pixel is marked as background. This is a feasible solution from an energy perspective because the active pixels in the score maps are so sparse (only 1 pixel is marked as key-point per score map in maximum) that their weights play a very small role in the loss function compared to the background pixels. To solve the two problems mentioned above, the original loss function is modified as follows. First, larger cost is assigned when the network classifies a keypoint pixel into background class; this helps the model stay away from local minima. Second, costs are assigned to the labeled pixels according to the distance between the labeled pixels and other key-points, which makes the model aware of distances. This loss function can be formulated as follows:

$$\sum_x \sum_y m(x,y) w \sum_k t_k(x,y) log(\frac{e^{\psi_k(x,y)}}{\sum_{k'} e^{\psi_{k'}(x,y)}}) \tag{7}$$

$$w = \begin{cases} \alpha, & k \in \{1 : KL\} \quad (7a) \\ \beta \cdot log(d((x,y),(x',y')) + 1), & k = KL + 1 \quad (7b) \end{cases}$$

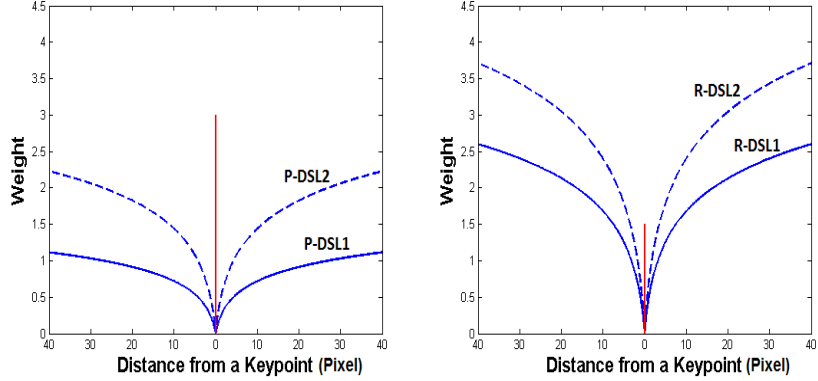

Figure 2: The shape of the distance-aware softmax loss (DSL) employed in the decision pathway. The transformations of the functions after increasing the values of $\beta$ are visualized through the dashed lines. The straight red lines indicate the cost of missclassifying a key-point pixel to a background pixel, while the blue lines indicate the cost of missclassifying a background pixel to a key-point pixel. (L) DSL for proposal, (R) DSL for refinement.

where $(x, y)$ are locations, $k \in \{1 : KL + 1\}$ is the index of classes, $\boldsymbol{\psi}_k(x, y)$ is the pixel value at $(x, y)$ in the $k^{th}$ score map of $\boldsymbol{\Psi}$, $t_k(x, y) = 1$ if $(x, y)$ belongs to class $k$, and 0 otherwise. The binary mask $m(x, y)$ is used to balance the amount of key-point and background pixels sampled in training. Different sampling ratios are employed for the background pixels that are nearby or further from the key-point. The threshold for differentiating 'nearby' and 'far-away' is measured by pixel distance on the score maps. In this paper, three pixels are employed as the threshold. The weight $w$ controls the penalty of foreground and background pixel. For a foreground pixel, a constant weight $\alpha$ is assigned to $w$, whose penalty is substantially larger than nearby background pixels. While for a background pixel, the distance $d((x, y), (x', y'))$ between the current pixel $(x, y)$ and a landmark location $(x', y')$ whose probability ranked the highest among the $KL$ foreground classes of the pixel is taken into account. As a result, the loss function assigns the weights based on the distance between the current pixel and the most misleading foreground pixel among the score maps, which punishes false-alarms adaptively. In Eq. (7b), a log function (base 10) is used to transform the distance into a weight, and employed a constant $\beta$ to control the magnitude of the cost. The shape of $w$ is depicted in Fig. 2. As a result, discrimination between the background and foreground pixels is encouraged according to the distance between a labeled pixel and a specific key-point. From a point of view of optimization, in back-propagation, since $d((x, y), (x', y'))$ is independent from $\psi_k(x, y)$, $w$ will be a constant which can be directly computed through Eq. (7a) or Eq. (7b).

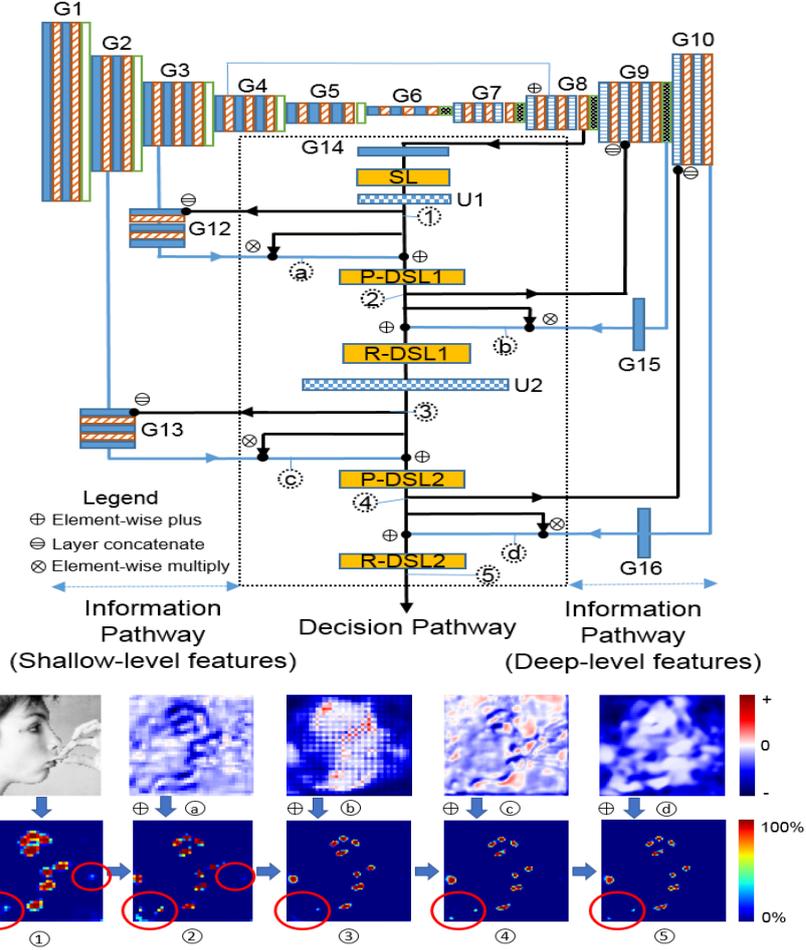

Figure 3: Depicted is the architecture of the proposed globally optimized dual-pathway (GoDP) model. Based on a naive DeconvNet [31], a precise key-point detector is derived by discriminatively extracting spatial and semantic information from shallow and deep layers of the network. The framework is motivated by cascaded regression, which contains residual error corrections and error feedback loops. Moreover, GoDP is end-to-end trainable, fully convolutional, and optimized from a pixel labeling perspective instead of traditional regression. Under the architecture, the $(KL + 1)^{th}$ score maps are visualized through sampling the network layers. The letter and number below each score map indicate the corresponding position in the network architecture. The background regions with red circles are highlighted to indicate how the proposal and refinement technique deal with background noises (best viewed in color).

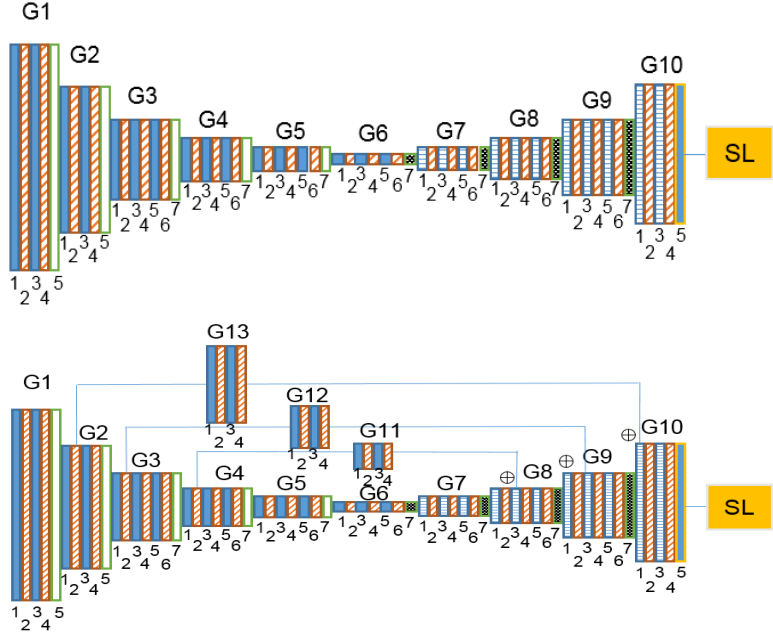

Figure 4: Depicted are the baseline network settings: (T) DeconvNet, (B) DeconvNet with Hour-glass [34] connections.

When training the network, Eq. (7b) is first replaced with a constant term (represented as $\beta$, which is less than $\alpha$). The network is trained with this degraded DSL (represented as SL in Fig. 3) for the first six epochs. Then, Eq. (7b) is employed for further refinement. During training, inspired by curriculum learning [41], the value of $\beta$ is gradually increased to encourage the network to discriminate pixels closer to the key-point locations.

### 3.5. Proposal and refinement in dual-pathway

To better exploit the spatial and semantic information encoded in a deep network, a dual pathway architecture is proposed as shown in Fig. 3. GoDP network is designed based on DeconvNet [31]. It is a classical architecture widely used for pixel-wise labeling. The intuition of the network can be explained as a supervised auto-encoder, which employs an encoder to extract high-level representation from the original image, then transforms the representation to a target semantic domain through a decoder. Recently a variation has been employed for face alignment [16, 32]. To alleviate spatial information loss caused by max-pooling layers, in DeconvNet, unpooling is used for preserving the stimulus structure before deconvolution. Derived from DeconvNet [31], the unique design of GoDP architecture includes separate pathways used for generating discriminative features and making decisions. They are designated as "information pathway" and "decision pathway". In the decision pathway, the depth of each layer is strictly kept as $KL + 1$ where each channel corresponds to a score map $\psi_k$. In the information pathway, depths of layers are unconstrained to enrich task-relevant information.

**Features in the information pathway:** The design of the information pathway is built upon the findings that feature maps generated from the deep layers of the network contain robust information that is invariant to the changing of image conditions but lack enough resolution to encode exact key-point locations. While the feature maps of shallow layers contain enough spatial information to localize the exact position of each key-point, they also contain a large amount of irrelevant noise. To handle this dilemma, a structure is built such that the features extracted from shallow layers are used to propose candidate regions while the features extracted from deep layers help to filter out false alarms and provide structural constraints. This is accomplished by imposing different losses to supervise shallow-level and deep-level features generated from shallow and deep layers. The parameters of DSL in the decision pathway are adjusted to enforce a large penalty when the shallow-level features fail to assign large positive probabilities to key-point locations, but give a smaller cost when they misidentify a background into a key-point candidate. This is a high detection rate policy to supervise shallow-level features. In contrast, a low false alarm policy is adopted to supervise deep-level features: high penalty is enforced when deep-level features misidentity a background pixel as key-point but slightly tolerate the error in the other way around. The results are shown in Fig. 3. After each shallow-level proposal, the contrast between background and foreground is increased, while, after each deep-level refinement, the background noise is suppressed. As a result, the key-point regions are gradually shrunk and highlighted.

**Score maps in the decision pathway:** In the decision pathway, the tensor $\mathbf{\Psi}^0$ is first initialized with the output of $2^{nd}$ deconvolution layers, where high-level information is well-preserved. Then, the probabilistic corrections $\triangle \mathbf{\Psi}^{t-1}$ generated from the shallow-level and deep-level layers of the network are computed and added to the decision pathway with the supervision of multiple DSLs. In testing, the DSLs work as conventional soft-max layers and can be seen as 'attention modules' on the information pathway, which helps the network re-focus on the target pixel locations. As shown in Fig. 3, during inference, score maps are first initialized on the decision pathway through Eq. (3), then concatenated with the layers in the information pathway through Eq. (4). These newly formed features are processed and aggregated into the decision pathway using Eq. (5), and finally the score maps in the decision pathway are updated by Eq. (6). The same process repeats several times to generate the final score maps. The intention of this architecture is identical to cascaded regression, where, in each stage, features are generated and contribute to reduce residual errors between predicted key-point locations and ground-truth locations. The predicted locations are updated and are used to re-initialize a new stage. The difference is that GoDP fully exploits the information encoded in a single network instead of resorting to a stacked architecture.

**Network structure**: In Fig. 3, a standard DeconvNet architecture containing ten groups of layers (G1, G2, G3, G4, G5, G6, G7, G8, G9, G10) is employed as feature source. Each group contains two or three convolutional/deconvolutional layers, batch normalization layers, and one pooling/unpooling layer. A hyperlink is added to connect G4 and G8 to avoid information bottleneck. The decision pathway is derived from the layer of G8, before unpooling. Bilinear upsampling layers are denoted as U1 and U2. Loss layer SL represents a degraded DSL (introduced in Section 3.3, represented as SL). P-DSL is used to represent DSL used for supervising key-point candidate proposal and use R-DSL to represent DSL used for supervising candidate refinement. Shapes of these DSLs are plotted in Fig. 2. The layers G12, G13, G14, G15, and G16 are additional groups of layers used to convert feature maps from information pathway to score maps in the decision pathway. The layers G12 and G13 contain three convolutional and two batch normalization layers. The layers G14, G15, and G16 include one convolutional layer. The settings of convolutional layers in G12 and G13 are the same: width 3, height 3, stride 1, pad 1 except the converters (last layer of G12 and G13) which connect the information pathway to the decision pathway, whose kernel size is $1 \times 1$. The other converters G14, G15, and G16 have the same kernel size: $1 \times 1$. The whole network takes 1GB GPU memory for a single image.

## 4. Experiments

In the experiments, the network is trained from scratch. For each score map, there is only one pixel at most that is marked as key-point pixel (depending on the visibility of the key-point). At the beginning, the network is trained with features generated from shallow-level layers only, which means the network has three loss functions instead of five in the first three epochs. After training the network for three epochs, the network is fine-tuned with all five loss functions for another three epochs. In the first six epochs, degraded DSLs (denotated as SL in Tables 2-6) are employed as the loss functions, then DSLs are employed so that the whole architecture is as shown in Fig. 3. The learning rate is gradually reduced from $10^{-3}$ to $10^{-7}$ during the whole training process. The stochastic gradient descent method (SGD) is employed to train the network. The input size of the network is $160 \times 160$ (gray scale) and the output size of score map is $80 \times 80$. It takes three days to train on one NVIDIA Titan X. The detailed parameter settings in training are shown in Tables 2-6.

### 4.1. Databases and baselines

Four highly challenging databases were used for evaluation: AFLW [1], AFW [27], UHDB31 [42], and 300-W [43]. The detailed experimental settings are summarized in Table 1. For evaluating on the AFLW database, the training and testing partition is strictly preserved as in [12]. Experiments are conducted on AFLW-PIFA (3,901 images for training, 1,299 images for testing, 21 landmarks annotated in each image) and ALFW-Full (20,000 training, 4,386 testing, 19 landmarks annotated in each image). The models trained on AFLW-PIFA are denotated as $M1$, and the models trained on AFLW-Full as $M2$. For evaluating on AFW database (468 images for testing, six landmarks annotated in each image), $M2$ is used. Six corresponding landmarks out of 19 are employed to report the performance. To evaluate the accuracy of algorithms under frontal faces, $M2$ is also employed. Different from the protocal employed in [12], all 1,314 images out of 4,386 in AFLW-Full database with 19 visible landmarks are considered as frontal faces and used for testing. Results are shown in Table 9 with the name AFLW-F. The database UHDB31 is a lab-environment database which contains 1,617 images, 77 subjects, and 12 annotated landmarks for each image. This is a challenging database including 21 head poses, combining seven yaw variations: [-90°:+30°:90°] and three pitch variations: [-30°:+30°:30°]. Nine landmarks (ID: 7,9,10,12,14, 15,16, 18,20 in [1]) are employed to compute landmark errors. Model $M2$ is employed for evaluating. The 300W is a well-studied medium pose [2] database which contains 68 annotated landmarks, including the points on facial contours. Following [25], the GoDP model is trained on the training set of Helen [44], LFPW [45] and the whole AFW [27]. The testing set of Helen, LFPW, and the challenging iBUG database [43] are used for evaluate. Same as Zhu *et al.* [25], the testing set of 300W is divided into the 'all', 'common', and 'challenging' subsets for comprehensive evaluating.

Table 1: Depicted are the detailed experimental settings.

| Evaluation Name | Training Set | # of Training Samples | Trained Model | Testing Set | # of Testing Samples | Point | Normalization Factor | Settings |
|---|---|---|---|---|---|---|---|---|
| AFLW-PIFA | AFLW | 3,901 | *M1* | AFLW | 1,299 | 21 | Face Size | Following [12] |
| AFLW-Full | AFLW | 20,000 | *M2* | AFLW | 4,386 | 19 | Face Size | Following [12] |
| AFLW-F | AFLW | 20,000 | *M2* | AFLW | 1,314 | 19 | Face Size | Section 4.1 |
| AFW | AFLW | 20,000 | *M2* | AFW | 468 | 6 out of 19 | Face Size | Section 4.1 |
| 300W-68 | 300W-68 | 3,148 | *M3* | 300W-68 | 689 | 68 | IOD | Section 4.1 |
| 300W-49 | 300W-68 | 3,148 | *M3* | 300W-49 | 689 | 49 out of 68 | IOD | Section 4.1 |
| 300W-13 | 300W-68 | 3,148 | *M3* | 300W-13 | 689 | 13 out of 68 | IOD | Section 4.1 |
| UHDB31 | AFLW | 20,000 | *M2* | UHDB31 | 1,617 | 9 out of 19 | Face Size | Section 4.1 |

Table 2: Depicted are the parameters of SL.

| Stage | Sampling ratio: Far-away pixels | Sampling ratio: Nearby pixels | Value of $\alpha$ | Value of $\beta$ | Type of Loss | Epoch |
|---|---|---|---|---|---|---|
| 1 | 0.005 | 0.1 | 1 | 0.2 | SL | 3 |
| 2 | 0.005 | 0.1 | 1 | 0.2 | SL | 3 |
| 3 | 0.005 | 0.1 | 1 | 0.2 | SL | 3 |

Table 3: Depicted are the parameters of P-DSL1.

| Stage | Sampling ratio: Far-away pixels | Sampling ratio: Nearby pixels | Value of $\alpha$ | Value of $\beta$ | Type of Loss | Epoch |
|---|---|---|---|---|---|---|
| 1 | 0.005 | 0.1 | 1 | 0.2 | SL | 3 |
| 2 | 0.001 | 0.2 | 3 | 0.1 | SL | 3 |
| 3 | 0.001 | 0.15 | 3 | 0.6 | DSL | 3 |

Table 4: Depicted are the parameters of R-DSL1.

| Stage | Sampling ratio: Far-away pixels | Sampling ratio: Nearby pixels | Value of $\alpha$ | Value of $\beta$ | Type of Loss | Epoch |
|---|---|---|---|---|---|---|
| 1 | - | - | - | - | - | - |
| 2 | 0.01 | 0.05 | 1 | 0.3 | SL | 3 |
| 3 | 0.01 | 0.05 | 1.5 | 1 | DSL | 3 |

Table 5: Depicted are the parameters of P-DSL2.

| Stage | Sampling ratio: Far-away pixels | Sampling ratio: Nearby pixels | Value of $\alpha$ | Value of $\beta$ | Type of Loss | Epoch |
|---|---|---|---|---|---|---|
| 1 | 0.005 | 0.1 | 1 | 0.2 | SL | 3 |
| 2 | 0.001 | 0.2 | 3 | 0.1 | SL | 3 |
| 3 | 0.001 | 0.15 | 3 | 0.6 | DSL | 3 |

Table 6: Depicted are the parameters of R-DSL2.

| Stage | Sampling ratio: Far-away pixels | Sampling ratio: Nearby pixels | Value of $\alpha$ | Value of $\beta$ | Type of Loss | Epoch |
|---|---|---|---|---|---|---|
| 1 | - | - | - | - | - | - |
| 2 | 0.01 | 0.05 | 1 | 0.3 | SL | 3 |
| 3 | 0.01 | 0.05 | 1.5 | 1 | DSL | 3 |

Multiple state-of-the-art methods (CDM [23], RCPR [21], CFSS [25], ERT [46], SDM [22], PO-CR [11], CCL [12], CALE [47], HF [33], PAWF [13], LBF [24], 3DDFA [2], MDM [48]) are selected as baselines. Normalized mean error (NME) is employed to measure the performance of algorithms as in [12]. As shown in Table 1, the bounding box size (face size) is employed to normalize landmark errors on AFLW, AFW, and UHDB31 database due to large head pose variations. Inter-ocular distance (IOD) is employed to normalize landmark errors on 300-W database.

For training and testing GoDP on AFLW database, the bounding box provided by [12] is employed. When the bounding box is not available or is not a rectangle (*e.g.*, on UHDB31 database), the bounding box generator provided by the authors of AFLW is employed to generate a new bounding box based on the visible landmarks and use it to initialize the methods and compute the NME. To preserve all the baselines' performance, the bounding boxes provided by the original authors are used for initialization. If the initialized bounding boxes are not available, for fair comparison, the bounding boxes are carefully initialized to maximize each baseline's performance.

### 4.2. Architecture analysis and ablation study

The objective of this experiment is to measure the impact of each proposed module in GoDP to the performance of the whole architecture. Along with the development of deep learning, the network structures become very complex, which makes the functionality of individual module unclear. To evaluate different networks' capability for generating discriminative score maps, new connections/structures of recent architectures are analyzed on the DeconvNet platform [31] to control uncertainty. The hourglass network (HGN) [34] is a recent extension of DeconvNet. The core contribution of Hour-glass net is that it aggregates features from shallow to deep layers through hyper-connections, which blends the spatial and semantic information for discriminative localization. Without GoDP's supervised proposal and refinement architecture, the information fusion of HGN is conducted in an unsupervised manner. The implementation of Hour-glass net is based on DeconvNet, three hyper-links are added to connect shallow and deep layers but residual connections are removed [49]. This model is selected to be one of the baselines. The detailed network settings for the model implementation can be viewed in Fig. 4.

Table 7: Depicted is the performance of different network architectures on PIFA-AFLW database. MPK (%) represents mean probability of key-point candidate pixels (large is better). MPB (%) represents mean probability of background pixels (small is better). NME-Vis represents NME (%) of visible landmarks. NME-All represents NME (%) of all 21 landmarks.

| Method | MPK | MPB | NME-Vis | NME-All |
|--------|------|------|---------|---------|
| DeconvNet [31] | 51.83 | 4.50 | 4.13 | **8.36** |
| HGN [34] | 28.38 | 0.96 | 3.04 | 11.05 |
| GoDP−DSL−PR | 31.79 | 1.01 | 3.35 | 13.30 |
| GoDP−DSL | 26.15 | 0.86 | 3.87 | 13.20 |
| GoDP | 39.78 | 1.30 | **2.94** | 11.17 |
| HGN(A) | 48.42 | 14.19 | 3.08 | 5.04 |
| GoDP(A) | 47.59 | 6.74 | **2.86** | **4.61** |

Firstly, the network is trained to detect visible landmarks. To evaluate the performance, a mask is employed to separate foreground and background pixels as shown in Fig. 1. Then the mean probability of foreground and background pixels are computed based on the mask. The mean probabilities are averaged over all testing images on the PIFA-AFLW database and the numbers are denotated as MPK and MPB in Table 7. These numbers indicate the contrast between the keypoint locations and background regions on the generated score maps. It is observed that GoDP performs significantly better in discriminating foreground and background pixels than other baseline structures and has a smaller landmark detection error. In ablation study, an architecture that removed DSL (shows as '-DSL' in Table 7) and another architecture that removed both DSL and proposal-refinement (PR) architecture (degraded DSL everywhere with the same parameter settings) are evaluated. The results are as shown in Table 7 as 'GoDP-DSL-PR'. These results indicate that DSL is critical for training a highly discriminative key-point detector and also contributes to regularize the proposal-refinement architecture. Additionally, it is observed that HGN outperforms DeconvNet. An explaination is the hyper-links introduced in HGN suppress background noise in DeconvNet.

Next, GoDP is trained to detect occluded landmarks. In the last stage of training (stage 3 in Table 2-6), instead of using the visible landmarks to train the network, the coordinates of all landmarks are used to train the last two DSLs (previous DSL/SL are trained with visible landmarks). The results are shown in Table 7 with the name **GoDP(A)**. HGN is also trained to detect occluded landmarks in the same way, the result is as shown in HGN(A). It is observed that GoDP(A) performs better in detecting both visible and invisible landmarks if the network is trained in this manner, and yields much more discriminative score maps. From this point forward, GoDP(A) is employed in the experiments described in the following sections.

### 4.3. Configuration of subspace partition

The objective of this experiment is to measure the impact of parameter settings of 3D pose-aware score map to the final performance of the whole architecture. Since pose encodes multi-modality variations of facial appearance, several recent works divide the target space of landmark detection into $K$ subspaces according to pose. The effects of $K$ are analyzed on two databases. The results are shown in Table 8. Three recent state-of-the-art methods are employed as baseline: CCL [12], DAC-CSR [50], and CALE [47]. The performance of GoDP(A) under $K$=1 and $K$=3 are tested ($K = 3$ is the default configuration in all the experiments). The results show that GoDP(A) behaved differently on the two databases. Note that on PIFA-AFLW, $K$=1 yields better performance, however, on AFLW-Full, $K$=3 works better. This phenomenon might be caused by the amount of training data. AFLW-Full includes more training samples than PIFA-AFLW and excludes two ambiguity points under both ears. These advantages help to train a stable and accurate model per subspace.

### 4.4. Performance on AFLW and AFW databases

The objective of this experiment is to compare GoDP(A) with state-of-the-art cascaded regression based key-point detectors on the challenging AFLW and AFW databases. They contain a large amount of faces with poses up to $90°$ yaw variations. In the implementation of this paper, Hyperface (HF) is trained without the loss of gender. The network architecture remains the same. The performance of 3DDFA and PAWF are reported based on the code provided by their authors. For the non-deep learning based methods, the performances from [12] are directly cited in this paper because the author's evaluating protocol (ID list of images, bounding boxes, and etc.) are strictly followed in the experiments.

Table 8: Depicted is the analysis of pose subspace partition on PIFA-AFLW and AFLW-Full database. Along with the results of GoDP(A), the most recent state-of-the-art methods are presented on these datasets. The number of partitions employed in the works is denoted by $K$. The NME (%) of all landmarks are reported in the databases.

| Method | Evaluation | K=1 | K=3 | K=5 | K=16 |
|---|---|---|---|---|---|
| CCL [12] | PIFA-AFLW | - | - | - | 5.81 |
| CALE-detector [47] | PIFA-AFLW | 5.53 | - | - | - |
| CALE-regressor [47] | PIFA-AFLW | 4.38 | - | - | - |
| GoDP(A) | PIFA-AFLW | **4.33** | 4.61 | - | - |
| CCL [12] | AFLW-Full | 3.73 | - | - | 2.72 |
| DAC-CSR [50] | AFLW-Full | - | - | 2.08 | - |
| GoDP(A) | AFLW-Full | 1.93 | **1.84** | - | - |

First the performance of four deep learning based methods in detecting **visible** landmarks are compared on challenging databases. Table 9 indicates that GoDP(A) performs the best, HF ranked second. Because HF relies on a global mapping from image ROI to shape space, it is not as discriminative as GoDP(A) in terms of localizing exact key-point positions. Deep cascaded regression methods like PAWF and 3DDFA performed well but not as accurately as GoDP(A). This observation can be further demonstrated by detecting landmarks on frontal faces. On the AFLW-F database, GoDP(A) performed significantly better than HF and 3DDFA on Table 9. The results indicate that GoDP(A) is more precise in localizing visible landmarks in both frontal and challenging poses.

Table 9: Depicted is the NME (%) of visible landmarks on multiple database partitions. The deep learning based methods are mainly compared in this table.

| | Deep Cascaded R. | | Deep End-to-End | |
|---|---|---|---|---|
| Evaluation | PAWF | 3DDFA | HF | GoDP(A) |
| AFLW-PIFA | 4.04 | 5.42 | - | **2.86** |
| AFLW-Full | - | 4.52 | 3.60 | **1.64** |
| AFLW-F | - | 4.13 | 2.98 | **1.48** |
| AFW | 4.13 | 3.41 | 3.74 | **2.12** |

Table 10: Depicted is the NME (%) of all annotated landmarks on AFLW-PIFA (3,901 training images) and AFLW-Full (20,000 training images) database.

| Baseline | Non-Deep Learning methods | | | | | | | | | Deep Cascaded R. | | Deep End-to-End | |
|---|---|---|---|---|---|---|---|---|---|---|---|---|---|
| | CDM | RCPR | CFSS | SDM | ERT | PO-CR | LBF | CCL | DAC-CSR | PAWF | 3DDFA | HF | GoDP(A) |
| AFLW-PIFA | 8.59 | 7.15 | 6.75 | 6.96 | 7.03 | - | 7.06 | 5.81 | - | 6.00 | 6.38 | - | **4.61** |
| AFLW-Full | 5.43 | 3.73 | 3.92 | 4.05 | 4.35 | 5.32 | 4.25 | 2.72 | 2.08 | - | 4.82 | 4.26 | **1.84** |

Then the performance of the methods in localization of **all** (both **visible** and **occluded**) landmarks are compared. The results are shown in Table 10. GoDP(A) performs consistently well on this experimental setting. The table also indicates that the performance of non-deep learning based methods are significantly improved when training database is enlarged, however, consistently underperforms GoDP(A). Compared to the non-deep learning based methods, the advantages of (*e.g.*, PAWF and 3DDFA) does not hold anymore, which may be caused by the indirect mapping from 3D pose parameters to 2D landmarks. The CED curves of deep learning based methods are plotted in Fig. 5. More qualitative results are presented in Figs. 11 and 12.

To further review the properties of GoDP, the robustness of GoDP(A) and HF (regression-based method) are compared under different bounding box initializations. This is important because bounding boxes generated by real face detectors always vary in size and position. Gaussian noises are added to the provided bounding boxes of AFLW-Full. The noise is generated based on the size of bounding boxes, where $\sigma$ controls the intensity of the Gaussian noise. The noise is added on the size and location of bounding boxes, and the results are as shown in Fig. 6, which discloses that GoDP(A) is much more robust to the initialization of bounding boxes than HF. It also discloses that GoDP(A) is more robust to variations of bounding box than translation variations. One explanation is that because GoDP(A) is a detection-based method, it is unable to predict any key-points outside the response region, but regression based methods can. One solution to compensate for this limitation is through randomly initializing multiple bounding boxes as ERT [46] and predicting landmark locations using median values. A more detailed comparison of GoDP(A) and ERT under automatic face detectors will is presented in Section 4.6.

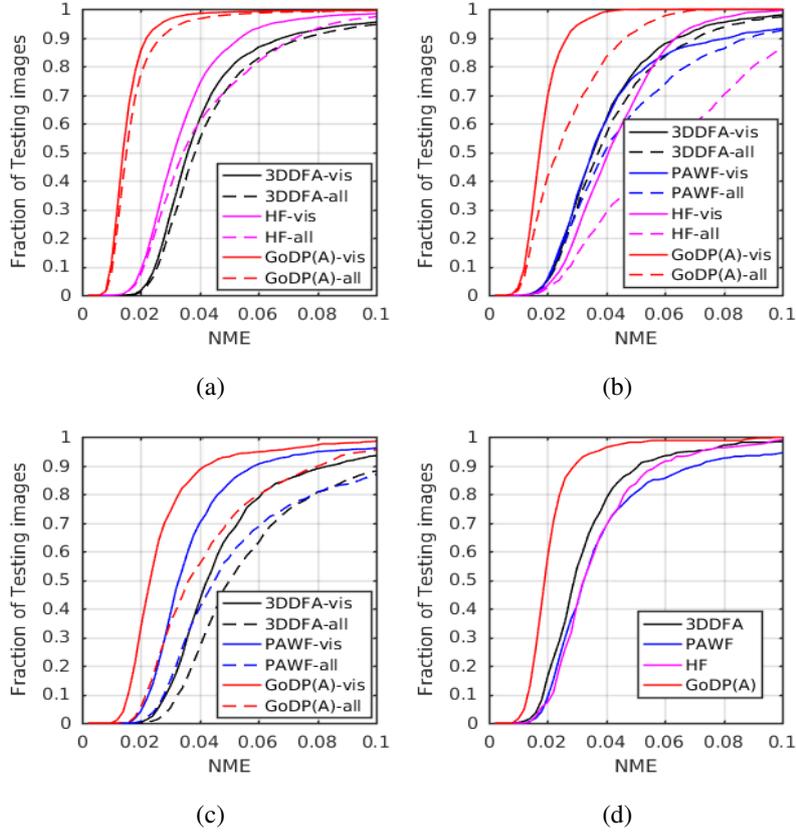

Figure 5: Depicted are the CED curves of deep learning based methods. 'vis'/'all' represents the error of visible/all landmarks. (a) AFLW-Full: 4,386 images, 19 landmarks. (b) UHDB31: 1,617 images, 9 landmarks. (c) AFLW-PIFA: 1,299 images, 21 landmarks. (d) AFW: 468 images, 6 landmarks. GoDP outperforms the other methods in all databases.

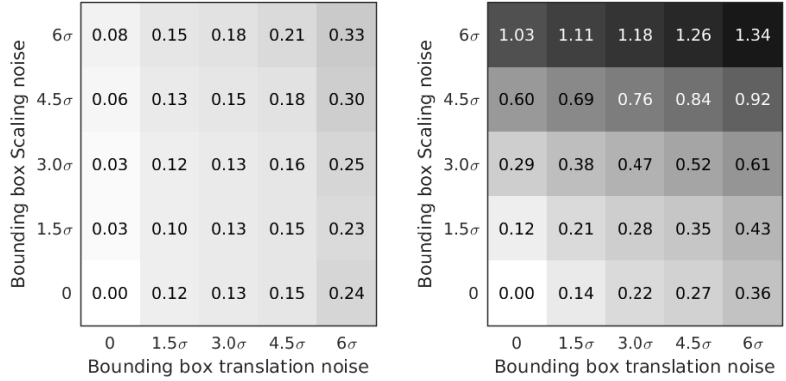

Figure 6: Depicted are the results of robustness evaluation. NME (%) increasing on all 19 landmarks of AFLW-Full database 4,386 images under noised bounding box initializations. The $\sigma$ is measured in percentage. (L) GoDP(A). (R) HF.

### 4.5. Performance on 300W database

The objective of this experiment is to further study the properties of GoDP on the well-studied 300W database [43]. GoDP is compared with multiple other methods as their implementations are publicly available. The input bounding boxes are either provided by the original authors (*e.g.*, MDM and CFSS) or carefully initialized from the ground-truth landmarks to maximum its performance (*e.g.*, ERT and SDM). The CED curves are shown in Fig. 7. The algorithms are evaluated to detect 68 or 49 landmarks (without facial contour [22]). It is observed that GoDP performs close to CFSS, outperforms SDM, ERT, and a recent deep recurrent network implementation: MDM in all experimental partitions. One explanation of why GoDP underperforms CFSS on 300W database (especially on the challenging partition), but outperforms it on the AFLW database is because: (1) the number of images in 300W training data is not enough to train an accurate GoDP model. The partition of iBUG database contains poses that have not been comprehensively covered in training. As a known limitation of deep learning based method, GoDP could not generalize well to these unseen poses as CFSS. This problem may also impact MDM. (2) Different from CFSS, GoDP lacks of global shape constraints. This constraint helps to localize non-fiducial landmarks (*e.g.*,

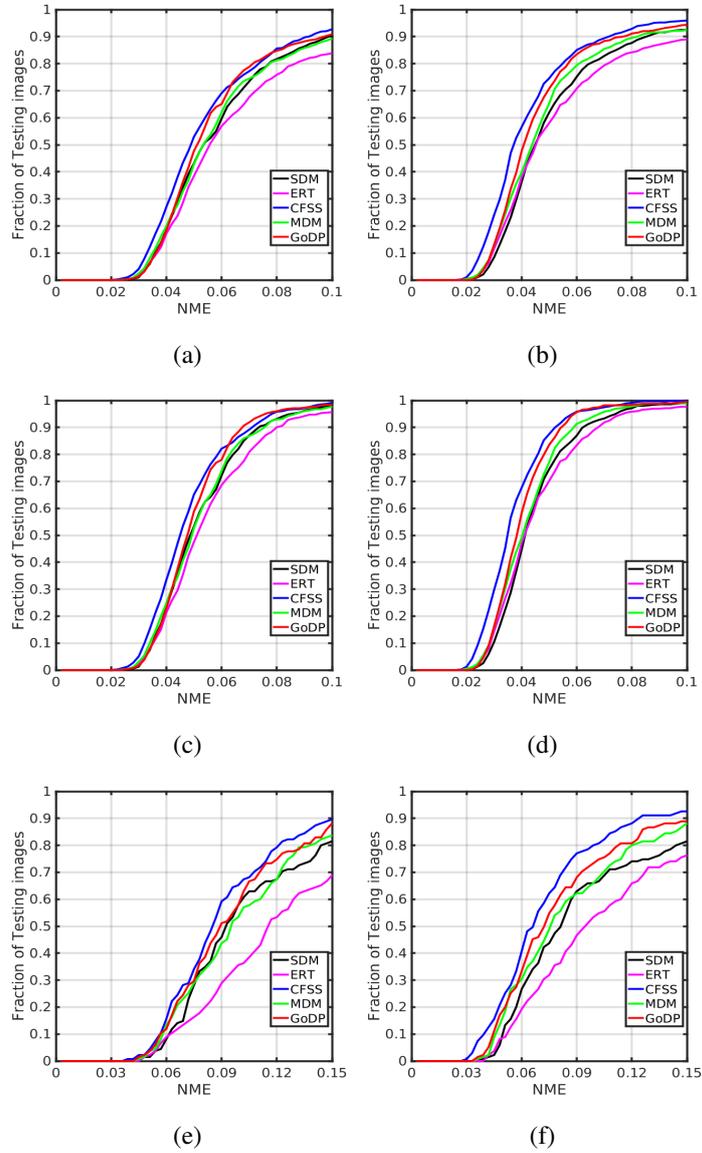

Figure 7: Depicted are the CED curves on 300W database. (a) 68 landmarks, all testing images. (b) 49 landmarks, all testing images. (c) 68 landmarks, common partition. (d) 49 landmarks, common partition. (e) 68 landmarks, challenging partition. (f) 49 landmarks, challenging partition.

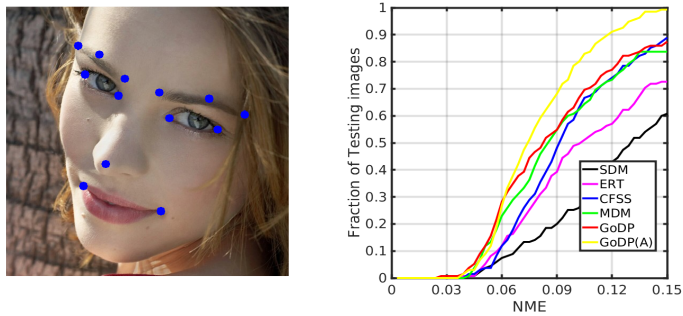

Figure 8: (L) Depicted are the positions of the 13 landmarks on a facial image. (R) Depicted are the CED curves of 13 landmarks on challenging partition.

19 landmarks on facial contours) in a consistent and accurate manner without resorting to large amount of training samples. In summary, experiment on 300W indicates that a large database is essential for training an accurate GoDP model.

To demonstrate the above claims more clearly and disclose the strength of GoDP, the NME of 13 landmarks is reported on challenging partition in Fig 8(L). The positions of 13 landmarks are shown in Fig 8(L), they are the 13 fiducial landmarks that share the same semantic meaning between the 21 annotations of AFLW database and 49 annotations of iBUG database. Hence, GoDP(A) trained on AFLW-Full database can be used to detect the 13 landmarks on 300W database. In Fig 8(R), it is observed that GoDP, trained on 300W, outperforms CFSS. Meanwhile, GoDP(A), trained on AFLW database, significantly outperforms all the other methods. The experimental result indicates that GoDP is able to estimate the 13 fiducial points (*e.g.*, eye corners, mouth corners) more accurate than other non-fiducial landmarks included in the 68 or 49 landmark protocal. The experiment also discloses that GoDP(A), trained on a large scale database covers very large head pose variations, outperforms the model trained on a smaller database with limited head pose variations. Considering the results present in Table 10, it shows that GoDP(A) model has better generalization capability than CFSS when feed into more training data, which discloses the benefits of GoDP.

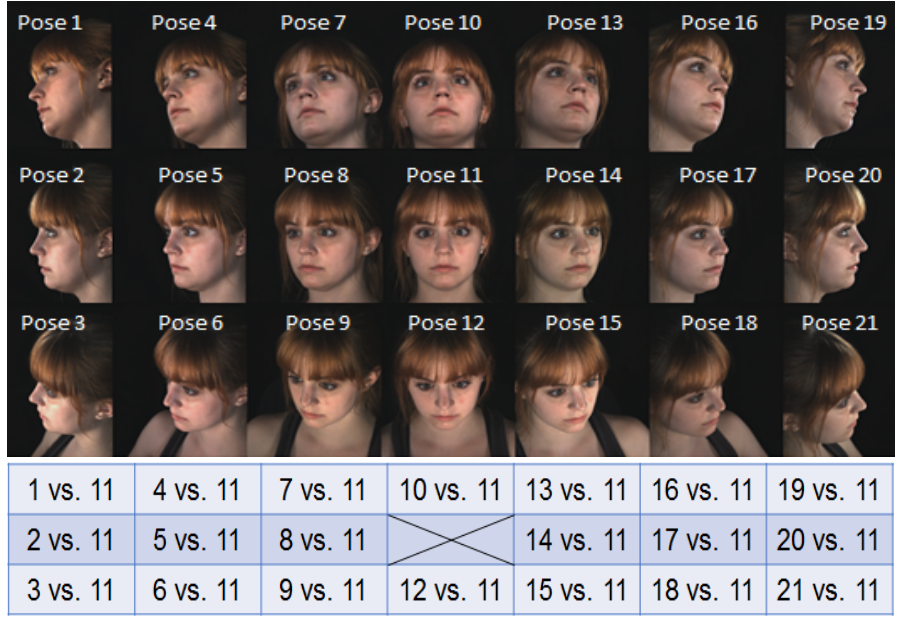

Figure 9: (T) Depicted are the 21 views of UHDB31 database. (B) In the depicted image, the center view is selected as gallery, the other 20 views are selected as probe. Each probe image is compared with all the 77 frontal gallery images in the database to determine its identity.

## 4.6. Performance on UHDB31 database

In the previous experiments, GoDP is demonstrated achieves state-of-the-art results in localizing fiducial landmarks under large head pose variations if the model under good initialization. It is believed the actual performance of a landamrk detector must be examined through a real application scenario where the data and experimental settings are both less controlled. The objective of this experiment is to disclose the performance of GoDP under a real face identification system where the landmark detector can hardly be initialized well. UHDB31 database, a new database used for evaluating pose-invariant face identification methods, is employed for benchmarking the performance of landmark detectors. The pose spectrum of this database is shown in Fig. 9. To illustrate the challenges of UHDB31 brings, Table 11 and Table 12 present the pose spectrum of several existing in-the-wild databases with the percentage of samples determined by yaw variations. UHDB31 database is more challenging compare to the other databases since 57.1% of total images have yaw variations larger than 36°, compare to 35.3% and 18.5% in AFLW and iBUG database. The difference is more significant if the number of samples with pitch variations equal or larger than 30° are considered. However, the iBUG and ALFW database do not have enough samples in this pose spectrum.

Table 11: Depicted is the percentage of samples (%) under different yaw variations in UHDB31 database.

| Database | $\geqslant 0°$ | $\geqslant 18°$ | $\geqslant 36°$ | $\geqslant 54°$ | $\geqslant 72°$ | $\geqslant 90°$ |
|----------|------|-------|-------|-------|-------|-------|
| iBUG [43] | **100** | 81.5 | 18.5 | 0 | 0 | 0 |
| AFLW [1] | **100** | 59.6 | 35.3 | 21.5 | 11.7 | 3.8 |
| UHDB31 | **100** | **85.7** | **57.1** | **57.1** | **28.6** | **28.6** |

Table 12: Depicted is the percentage of samples (%) under different yaw variations compound with the variations of pitch angle that equal or larger than 30° in UHDB31 database.

| Database | $\geqslant 0°$ | $\geqslant 18°$ | $\geqslant 36°$ | $\geqslant 54°$ | $\geqslant 72°$ | $\geqslant 90°$ |
|----------|------|-------|-------|-------|-------|-------|
| iBUG [43] | 5.9 | 2.2 | 0 | 0 | 0 | 0 |
| AFLW [1] | 5.0 | 2.9 | 1.7 | 1.1 | 0.6 | 0.3 |
| UHDB31 | **66.7** | **57.1** | **38.1** | **38.1** | **19.1** | **19.1** |

Because the database contains facial images with large pitch and yaw variations, the GoDP(A) model trained on AFLW-Full is employed to localize landmarks. To fully understand the performance of landmark detectors on this database, GoDP(A) and other baseline methods are firstly initialized from the ground-truth bounding boxes for benchmarking. The CED curve is shown in Fig. 5(b) and the quantitative comparison is shown in Fig. 13. These results show that GoDP(A) performs consistently well in UHDB31 across pose variations. Besides, Fig. 5(b) also indicates that the performance of GoDP(A) and HF is not as consistent as 3DDFA and PAWF in terms of detecting both visible and invisible landmarks. Because 3DDFA and PAWF rely on parametric 3D shape model to localize 2D landmarks, 3D shape provides a stronger geometric regularization over the invisible landmarks.

Next, the performance of GoDP(A) is analyzed from a systematic point of view. In real-world deployment, facial bounding box cannot be initialized from ground-truth landmarks, which strongly affects the performance of landmark detectors. Hence, in this experiment, GoDP is initialized from automatic face detectors. ERT is selected as baseline because it has became a defacto landmark localization module in recent facial analysis researches [5, 6, 7], thanks to the Dlib [4] toolbox. To initialize the bounding box for GoDP(A), faces are first detected based on DPM head hunter [3], which is able to detect more than 95% of faces on UHDB31 database. Then, a pre-trained deep bounding box regressor (similar to [51]) is employed to refine the facial ROIs to approximate the bounding boxes defined in AFLW-Full, the database used to train GoDP(A). The NME is computed on the images where face detector successfully detect the face. In Table 13, the performance of GoDP(A) and ERT are compared on 21 views of UHDB31 in terms of the mean NME in landmark detection. This result demonstrates that GoDP(A), coupling with an automatic face detector, is a more robust and accurate choice for localizing landmarks in all head pose variations than ERT (Dlib).

Finally, the performance of landmark detector is investigated in a fully automatic 3D-aided face identification pipeline proposed by Kakadiaris *et al.* [8, 9, 10]: UR2D. UR2D uses 3D models to transform facial textures into canonical UV spaces, where facial pose is normalized to frontal. An extension of the original pipeline proposed by Xiang *et al.* [52] is employed for face identification. Following the 1-vs-all experimental protocol [42] on UHDB31 database, 77 images (one image per subject) in frontal view are selected as gallery and all the other views of UHDB31 database (77 subjects $\times$ (21-1) poses = 1540 images in total) are selected as probe. A sample of this evaluation protocol is shown in Fig. 9. The cosine similarity scores between a probe image with all 77 gallery images are computed to determine the ID of the probe. Fig. 10 shows that GoDP, initialized from the bounding boxes provided by head-hunter, significantly improves the quality of face frontalization in all poses compared with using ERT (Dlib). Specifically, it improves the rank-1 identification performance by 44.2% in average. The results indicate that GoDP(A) is a better choice than ERT (Dlib), a widely used landmark estimator, in a fully automatic 3D-aided face identification system.

Table 13: Depicted is the performance of GoDP(A) / ERT(Dlib) on 21 views of UHDB31 under automatic face detectors. NME (%) of nine landmarks is reported (as shown in Fig. 13). Columns correspond to pitch variations, rows correspond to yaw variations.

| | -90° | -60° | -30° | 0° | 30° | 60° | 90° |
|---|---|---|---|---|---|---|---|
| 30° | **3.9**/12.9 | **3.1**/7.2 | **1.8**/3.7 | **1.7**/3.0 | **2.2**/4.1 | **3.5**/7.4 | **4.2**/10.5 |
| 0° | **3.8**/19.5 | **2.3**/4.8 | **1.7**/2.3 | **1.4**/2.2 | **1.8**/2.3 | **2.8**/4.2 | **4.1**/8.5 |
| -30° | **5.5**/- | **3.6**/6.8 | **2.0**/3.5 | **1.9**/3.0 | **2.1**/3.5 | **3.7**/6.7 | **5.2**/7.1 |

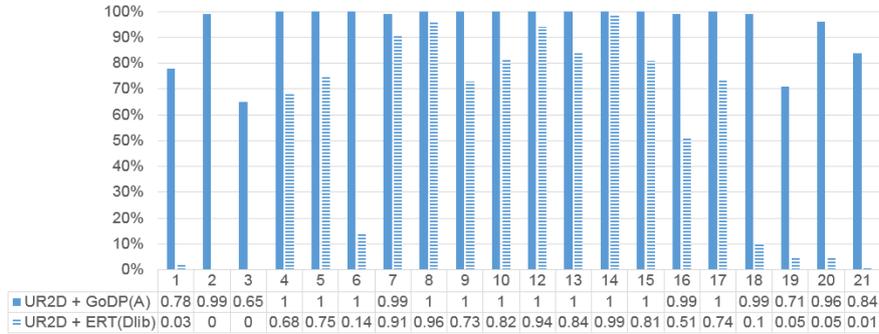

| | 1 | 2 | 3 | 4 | 5 | 6 | 7 | 8 | 9 | 10 | 11 | 12 | 13 | 14 | 15 | 16 | 17 | 18 | 19 | 20 | 21 |
|---|---|---|---|---|---|---|---|---|---|---|---|---|---|---|---|---|---|---|---|---|---|
| UR2D + GoDP(A) | 0.78 | 0.99 | 0.65 | 1 | 1 | 1 | 0.99 | 1 | 1 | 1 | 1 | 1 | 1 | 1 | 1 | 0.99 | 1 | 0.99 | 0.71 | 0.96 | 0.84 |
| UR2D + ERT(Dlib) | 0.03 | 0 | 0 | 0.68 | 0.75 | 0.14 | 0.91 | 0.96 | 0.73 | 0.82 | 0.94 | 0.84 | 0.99 | 0.91 | 0.81 | 0.51 | 0.74 | 0.1 | 0.05 | 0.05 | 0.01 |

Figure 10: Depicted is the rank-1 identification rate (%) of 20 poses on UHDB31 database.

## 5. Conclusion

In this paper, an efficient deep network architecture (GoDP) is proposed to localize facial landmarks with high precision. The architecture transforms the traditional regression problem into a 2D cascaded pixel labeling problem through a unique proposal-refinement technique and distance-aware loss functions. The experimental analysis demonstrated the precision and robustness of GoDP over multiple state-of-the-art deep and shallow structures on multiple challenging in-the-wild databases. GoDP, coupled with an automatic landmark detector, is demonstrated to be able to improve the overall performance of a 3D-aided face identification pipeline by 44.2% in average on a challenging database.

## 6. Acknowledgment

This material is based upon work supported by the U.S. Department of Homeland Security under Grant Award Number 2015-ST-061-BSH001. This grant is awarded to the Borders, Trade, and Immigration (BTI) Institute: A DHS Center of Excellence led by the University of Houston, and includes support for the project "Image and Video Person Identification in an Operational Environment: Phase I" awarded to the University of Houston. The views and conclusions contained in this document are those of the authors and should not be interpreted as necessarily representing the official policies, either expressed or implied, of the U.S. Department of Homeland Security.

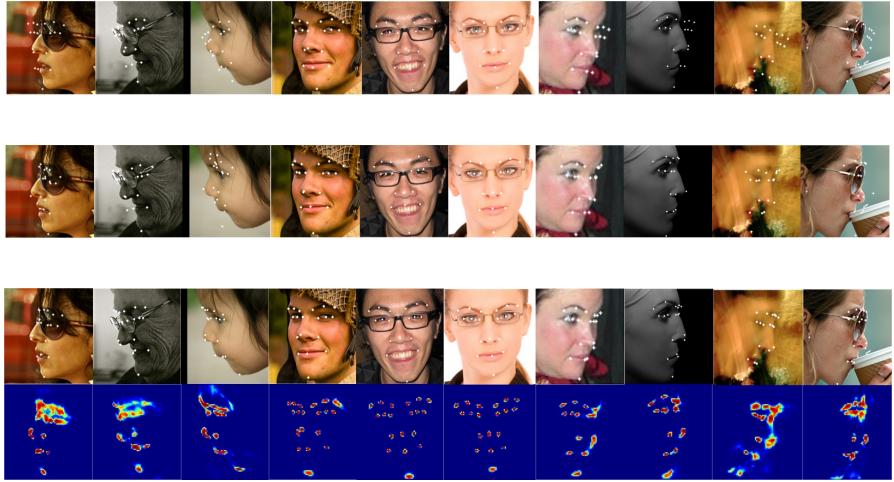

Figure 11: Depicted are the qualitative results on AFLW-Full database. (T) HF [33], (M) 3DDFA [2], (B) GoDP(A) with score maps.

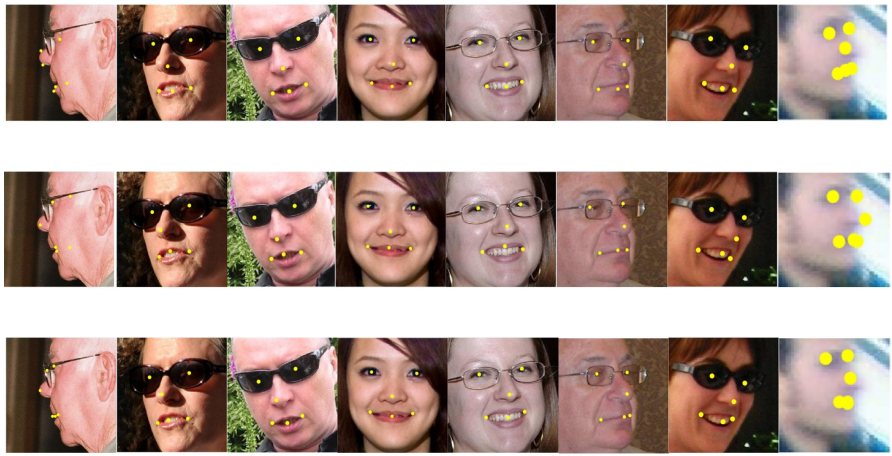

Figure 12: Depicted are the qualitative results on AFW database. GoDP(A) is able to localize key-points precisely. (T) HF [33], (M) 3DDFA [2], (B) GoDP(A).

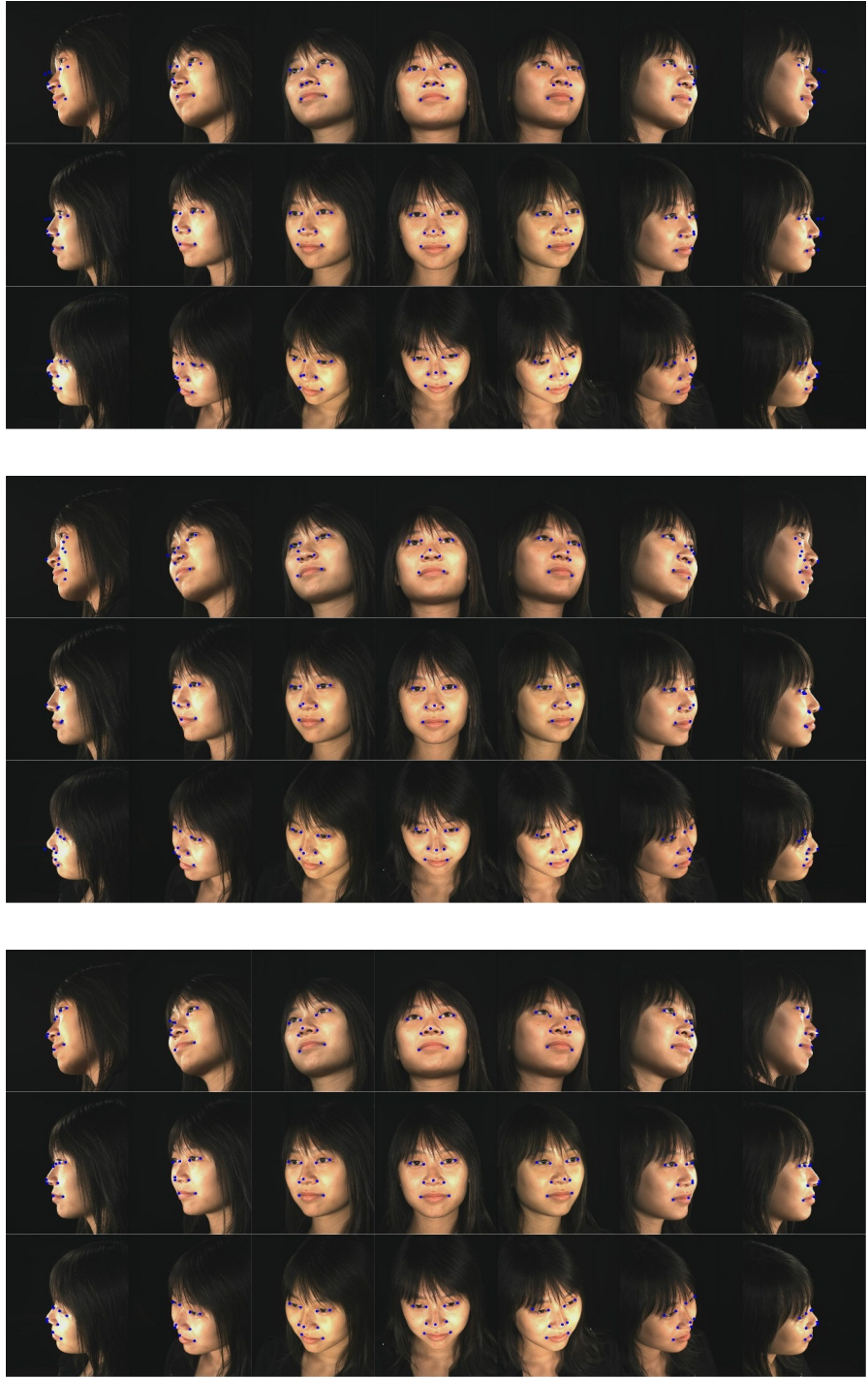

Figure 13: Depicted are the qualitative results on UHDB31 database. (T) HF [33], (M) 3DDFA [2], (B) GoDP(A).